\documentclass[10pt,conference,a4paper]{IEEEtran}
\usepackage{cite}
\usepackage{amsmath,amssymb,amsfonts}
\usepackage{algorithmic}
\usepackage{graphicx}
\usepackage{textcomp}
\usepackage{multirow}
\usepackage{tabularx}
\usepackage{subcaption} 
\usepackage{hyperref}
\usepackage{xcolor}
\def\BibTeX{{\rm B\kern-.05em{\sc i\kern-.025em b}\kern-.08em
    T\kern-.1667em\lower.7ex\hbox{E}\kern-.125emX}}

\newcolumntype{C}[1]{>{\centering\arraybackslash}p{#1}}

\makeatletter
\newcommand{\linebreakand}{%
  \end{@IEEEauthorhalign}
  \hfill\mbox{}\par
  \mbox{}\hfill\begin{@IEEEauthorhalign}
}

\begin{document}

\title{Local Clustering with Mean Teacher for Semi-supervised learning
}

\author{\IEEEauthorblockN{Zexi Chen}
\IEEEauthorblockA{
\textit{North Carolina State University}\\
Raleigh, USA \\
Email: zchen22@ncsu.edu}
\and
\IEEEauthorblockN{Benjamin Dutton}
\IEEEauthorblockA{
\textit{North Carolina State University} \\
Raleigh, USA \\
Email: bcdutton@ncsu.edu}\\
\and
\IEEEauthorblockN{Bharathkumar Ramachandra}
\IEEEauthorblockA{
\textit{North Carolina State University}\\
Raleigh, USA \\
Email: bramach2@ncsu.edu}
\linebreakand
\IEEEauthorblockN{Tianfu Wu}
\IEEEauthorblockA{
\textit{North Carolina State University}\\
Raleigh, USA \\
Email: tianfu\_wu@ncsu.edu}
\and
\IEEEauthorblockN{Ranga Raju Vatsavai}
\IEEEauthorblockA{
\textit{North Carolina State University}\\
Raleigh, USA \\
Email: rrvatsav@ncsu.edu}
}

\maketitle

\begin{abstract}
The Mean Teacher (MT) model of Tarvainen and Valpola has shown favorable performance on several semi-supervised benchmark datasets. MT maintains a teacher model's weights as the exponential moving average of a student model's weights and minimizes the divergence between their probability predictions under diverse perturbations of the inputs. However, MT is known to suffer from confirmation bias, that is, reinforcing incorrect teacher model predictions. In this work, we propose a simple yet effective method called Local Clustering (LC) to mitigate the effect of confirmation bias. In MT, each data point is considered independent of other points during training; however, data points are likely to be close to each other in feature space if they share similar features. Motivated by this, we cluster data points locally by minimizing the pairwise distance between neighboring data points in feature space. Combined with a standard classification cross-entropy objective on labeled data points, the misclassified unlabeled data points are pulled towards high-density regions of their correct class with the help of their neighbors, thus improving model performance. We demonstrate on semi-supervised benchmark datasets SVHN and CIFAR-10 that adding our LC loss to MT yields significant improvements compared to MT and performance comparable to the state of the art in semi-supervised learning. The code for our method is available at \url{https://github.com/jay1204/local_clustering_with_mt_for_ssl}.
\end{abstract}


\section{Introduction}

In recent years, deep neural networks have achieved great success in many supervised machine learning tasks such as image classification \cite{krizhevsky2012imagenet}, object detection \cite{he2017mask}, and video action recognition \cite{simonyan2014two}. However, these efforts involved training deep networks on large amounts of labeled data, and such successes have not followed when only a small amount of labeled data is available. Human annotations are costly to obtain and often pose their own unique challenges \cite{barriuso2012notes}. Herein lies the motivation for semi-supervised learning (SSL) techniques, which allow substantial amounts of cheaply available unlabeled data to supplement small amounts of expensive labeled data in training a machine learning model. 

Among the various approaches to SSL with deep networks, consistency-based methods~\cite{laine2016temporal,mean-teacher,virtual-adversarial} have set state of the art performance on multiple benchmark datasets~\cite{netzer2011reading,deng2009imagenet}. The primary goal of consistency-based methods is to encourage consistent probability predictions for the same data under either (1) different noise conditions or (2) different network parameterizations. In other words, consistency-based methods enforce smoothness around each data point locally in output space. This class of methods has roots in Ladder Networks~\cite{ladder-network}, a deep unsupervised learning method with an auto-encoder-like architecture. The model aims to learn abstract, invariant features in higher layers of the network that are robust to various kinds of noise added to either the input or intermediate representations. The $\Gamma$-model~\cite{rasmus2015semi} extends Ladder Networks to the semi-supervised setting by training the model using a multi-objective loss comprised of a supervised classification loss and an unsupervised consistency loss. 

Following $\Gamma$-model, the $\Pi$-model~\cite{laine2016temporal} and Mean Teacher model~\cite{mean-teacher} improve upon its performance by designing better quality teachers, which generate better learning objectives for their students. Taking a different perspective, VAT~\cite{virtual-adversarial} attempts to improve upon the $\Gamma$-model by selectively adding perturbations to inputs in \textit{adversarial} directions, ones that can most quickly get the student and teacher predictions to deviate from each other. 

\begin{figure}[t]
\centering
\mbox{
    \includegraphics[width=\linewidth]{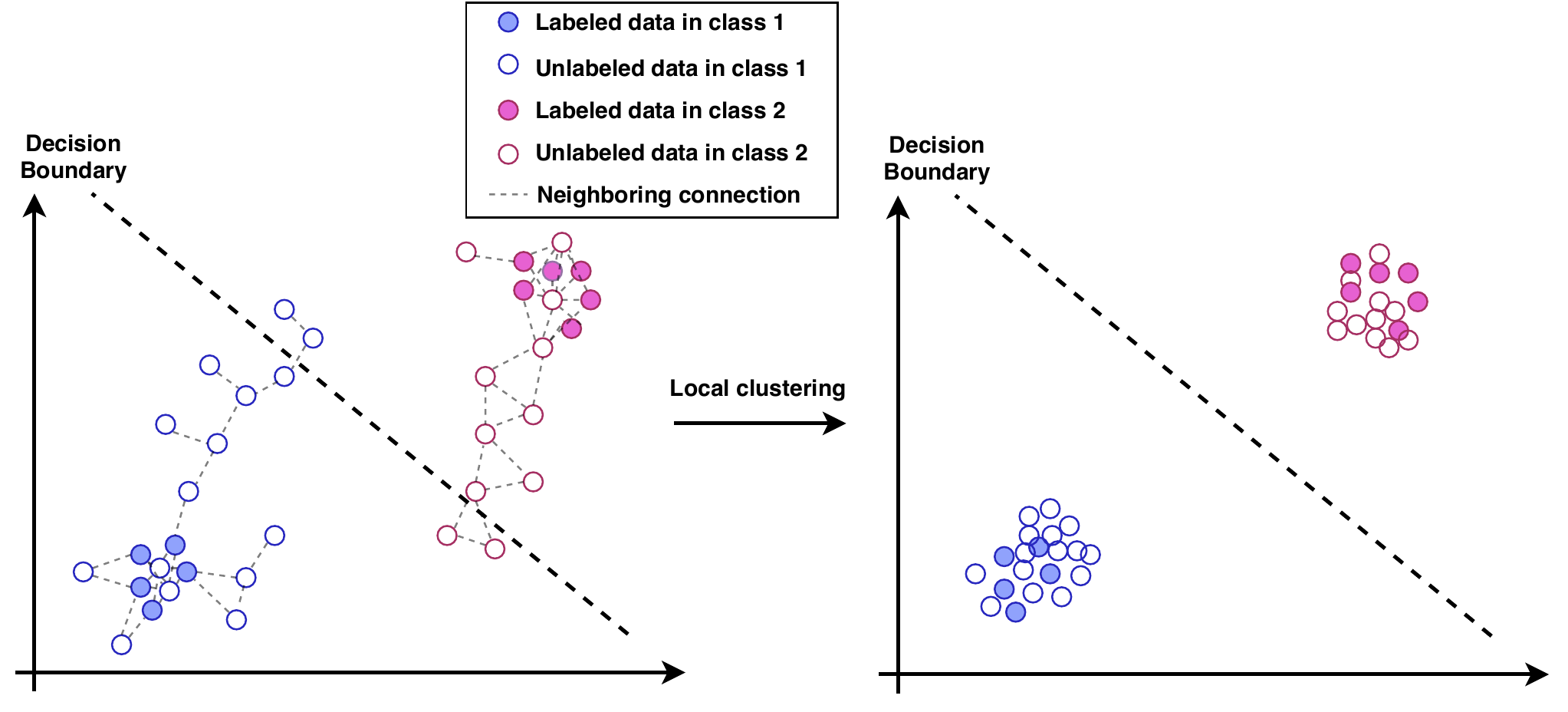}
  }
   \caption{An illustration of the intuition behind Local Clustering in feature space. Each point represents the intermediate learned representation of one data sample. Best viewed in color in electronic form.}
\label{fig:illustrate_idea}
\end{figure}

However, a well-known problem with these recent consistency-based methods is confirmation bias \cite{mean-teacher}. In requiring a student to be consistent with a teacher in terms of prediction, consistency-based methods assume that the teacher's predicted probabilities (targets for the student) are satisfactory. Confirmation bias in these approaches is caused by sub-optimal targets provided by the teacher model on unlabeled data, causing the student to reinforce these sub-optimal predictions. This problem is especially severe at the early stages of training, where the predictions from the teacher for many unlabeled data points can be inaccurate, causing those unlabel data points to be trapped in low-density regions in feature space, and those errors carry forward throughout the training. More importantly, the model is unable to learn discriminative features (class-specific knowledge) from those unlabeled data if they lie in low-density regions in feature space. 

In this work, we propose a local clustering method to address the above limitation. We choose Mean Teacher~\cite{mean-teacher} as our baseline and add a novel regularizer to it. Motivated by widely used \textit{local consistency assumption}\cite{zhou2004learning} in SSL, where nearby samples are likely to have the same label, we consider the vicinity of each unlabeled data sample in feature space and enhance the connections between neighboring samples. We propose the regularizer loss term that eventually pulls unlabeled data samples towards high-density regions with the assistance of neighboring samples. This is done by penalizing the distance between neighboring samples' feature representations. The intuition is that the weight updates thence caused would affect the learned features to help the model pick up on new cues from those unlabeled data samples, hopefully in the form of class-specific discriminative features. This intuition is illustrated in Figure \ref{fig:illustrate_idea}.

Empirically, our approach substantially improves accuracy over Mean Teacher and achieves comparable performance with state of the art methods on benchmark datasets SVHN and CIFAR-10 following identical neural network architecture and evaluation rules.

\section{Related Work}
\label{sec:related work}
Recent deep SSL methods can be broadly categorized into deep clustering methods, deep generative methods, and consistency-based methods. We provide a short review of the first two and a detailed review of the more relevant consistency-based methods.

\subsection{Deep Clustering methods}
In general, deep clustering methods consist of two stages, where the first stage is a feature extractor network for learning deep feature embeddings from input data and the second stage is a clustering algorithm such as k-means or graph-based clustering on extracted feature embeddings guided by partial labeled information. 
Hsu and Kira~\cite{hsu2015neural} proposed a method to train the two stages jointly in an end-to-end manner. Instead of employing class labels for supervision, they define weak labels in the form of similar/dissimilar data pairs, so that the clustering strategy is to minimize pairwise distances between similar pairs and maximize pairwise distances between dissimilar pairs. Following their work, Shukla $et$ $al.$~\cite{Shukla2018SemiSupervisedCW} proposed ClusterNet, an auto-encoder based framework composed of both the aforementioned pairwise constraint clustering and k-means based clustering, where the latter additionally penalizes high intra-cluster variance. Another recent approach \cite{haeusser2017learning} employs pairwise relations from a completely different perspective, where they define feature similarities as associations and form association cycles as a two-step walk starting and ending from labeled samples via an unlabeled sample in feature space. Accordingly, their objective is to encourage consistent association cycles that end at a sample of the same class as the starting sample and to penalize inconsistent ones. Apart from these methods, Kam-nitsas $et$ $al.$ \cite{kamnitsas2018semi} propose a compact latent space clustering approach in the spirit of graph-based approaches. They build the graph directly in feature space instead of obtaining a pre-constructed graph from input space as the more general graph-based approaches, and form single and compact clusters for each class by encouraging balanced intra-cluster transition probabilities and zero inter-cluster transition probabilities.  

\subsection{Deep Generative methods}
Deep generative methods have also been widely explored in recent years. Generative methods in SSL aim at estimating the joint distribution over data and class labels $p(\mathbf{x}, y)$, and compute a posterior distribution $p(y|\mathbf{x})$ via Bayes theorem. Two popular techniques with deep generative approaches in SSL are using variational auto-encoders (VAE)~\cite{kingma2013fast,rezende2014stochastic} and Generative Adversarial Networks (GANs)~\cite{goodfellow2014explaining}.

In VAE-based approaches\cite{kingma2014semi,rezende2015variational}, the authors propose semi-supervised learning using VAEs by assuming that the data $\mathbf{x}$ are generated from the latent class variable $y$ and a continuous latent variable $\mathbf{z}$. Consequently, the SSL problem is interpreted as a latent class missing variable problem. They construct a deep generative model and optimize it using variational inference to estimate the joint distribution $p(\mathbf{x}, \mathbf{z}, y)$ for labeled data and $p(\mathbf{x}, \mathbf{z})$ for unlabeled data. In these works, the continuous latent variable $\mathbf{z}$ is chosen as a diagonal Gaussian distribution. Maal{\o}e $et$ $al.$ \cite{maaloe2016auxiliary} further introduce an auxiliary variable which helps generalize $\mathbf{z}$ to a non-Gaussian distribution and thus improve model performance. 

In GANs~\cite{goodfellow2014explaining}, an adversarial game is set up between discriminator and generator networks. The objective of the generator is to generate fake samples that cannot be distinguished from real ones by the discriminator, which is tasked with telling them apart. Salimans $et$ $al.$ \cite{salimans2016improved} pioneered the extension of GANs to SSL. They propose a Feature Matching GAN (FM-GAN), where the objective of the generator is adjusted to match the first moment of the real data distribution in feature space, and the discriminator is a $(K + 1)$-head classifier with the extra class referring to fake samples from the generator. Following their work, Dai $et$ $al.$~\cite{dai2017good} propose a complement generator to address the limitations of the feature matching objective in FM-GAN. They theoretically show that a preferred generator for GAN-based SSL is to encourage generating diverse fake samples in low-density regions of the feature space, so that the real samples are pushed towards separable high-density regions and hence the discriminator is able to establish correct classification decision boundaries. The Localized GAN (LGAN) proposed by Qi $et$ $al.$~\cite{qi2018global} improves FM-GAN by making the discriminator resistant to local perturbations, where the perturbations are various fake samples produced by a local generator in the neighborhood of real samples on a data manifold.

Since our method builds on a consistency-based method, a detailed discussion for that category is provided in Section \ref{sec:consistency-based}.

\section{Preliminaries}
Given a general SSL problem, let $\mathcal{D}_l:={\{(x_i, y_i)\}}_{i=1}^{n_l}$ represent a set of labeled samples, and $\mathcal{D}_u:={\{x_i\}}_{i=1}^{n_u}$ represent a set of unlabeled data samples. Typically, the number of labeled samples is much smaller than the number of unlabeled samples, $n_l \ll n_u$. The objective of SSL is to learn a mapping function $f(\mathbf{x}; \theta): \mathcal{X} \rightarrow \mathcal{Y}$, from the input space $\mathcal{X}$ to the label space $\mathcal{Y}$, where $\mathcal{X} \in \mathbb{R}^d$, $\mathcal{Y} \in \{1, 2, ..., K\}$ and $K$ is the total number of classes. In our work, the mapping function $f(\mathbf{x}; \theta)$ is chosen to be represented with a deep neural network. We can further decompose this as $f(\mathbf{x}; \theta) = h(g(\mathbf{x}; \theta_g); \theta_h)$, where $\mathbf{z} = g(\mathbf{x}; \theta_g)$ is a feature extractor network mapping from input space $\mathcal{X}$ to latent space $\mathcal{Z}$, and $\mathbf{y} = h(\mathbf{z}; \theta_h)$ is a classification network mapping from latent space $\mathcal{Z}$ to label space $\mathcal{Y}$. Consistency-based methods~\cite{laine2016temporal,mean-teacher,virtual-adversarial} are the best performers among various classes of SSL methods and Mean Teacher is widely considered a state of the art baseline among consistency-based methods. We next briefly introduce consistency-based methods and the Mean Teacher model.

\subsection{Review of Consistency-based methods}
\label{sec:consistency-based}
Consistency-based methods, also called consistency regularizers, encourage consistent probability predictions under small changes to either the inputs or the parameters of the model. Typically, the perturbations are represented in the form of input augmentations, dropout regularization~\cite{srivastava2014dropout} or adversarial noise~\cite{goodfellow2014explaining}. Given two random perturbations $\xi'$ and $\xi''$ to input $x_i$, the general form of the consistency loss term can be formulated as the difference between a student model $f(\mathbf{x};\theta')$ and a teacher model $f(\mathbf{x};\theta'')$:
\begin{equation}
\label{eq1}
    \mathcal{L}_{cons} = \mathop{\mathbb{E}}_{{\{x_i\}}_{i=1}^{n_l + n_u}} D[f(x_i, \xi'; \theta'), f(x_i, \xi''; \theta'')]
\end{equation}
where $D[\mathord{\cdot}, \mathord{\cdot}]$ measures the difference between the probability predictions of $f(\mathbf{x};\theta')$ and $f(\mathbf{x};\theta'')$, usually chosen to be Mean Squared Error or KL divergence. A theoretical analysis of the consistency loss in \cite{athiwaratkun2018there} has shown that it improves the model generalization ability by penalizing the Jacobian norm and the Hessian eigenvalues of the predicted outputs with respect to inputs. It can be viewed as a regularization term leveraging both labeled and unlabeled data. The total loss for this class of methods integrates the consistency loss $\mathcal{L}_{cons}$ with the cross entropy loss $\mathcal{L}_{ce}$ defined on labeled samples, expressed as
\begin{equation}
    \mathcal{L} = \mathcal{L}_{ce} + \lambda \mathcal{L}_{cons}
\end{equation}
where the coefficient $\lambda$ is a hyperparameter that controls the relative importance of the consistency loss. In particular, several SSL approaches have been developed based on the idea of enforcing consistency including Mean Teacher. 

\hfill \break
\textbf{Mean Teacher}~\cite{mean-teacher}: In Mean Teacher, the weights $\theta''$ of the teacher model are maintained through training as the exponential moving average (EMA) of the weights of the student model, formulated as 
\begin{equation}
\theta_t'' = \alpha \theta'_{t-1} + (1 - \alpha) \theta_t'
\end{equation}
where $t$ indexes training iteration and the coefficient $\alpha$ is a smoothing hyperparameter. The main idea is to form a better teacher model, which gradually aggregates information from the student model in an EMA fashion. Eventually, a better teacher model can generate more stable probability predictions which serve as higher quality targets to guide the learning process of the student model. In \cite{mean-teacher}, the authors predicted that other methods which further improve the quality of targets would follow, and we believe that our local clustering is one such method.

\hfill \break
\textbf{Other consistency-based methods}: In $\Pi$ model~\cite{laine2016temporal}, the student model itself also serves as the teacher model, interpreted as $\theta' = \theta''$.
Since two random perturbations $\xi'$ and $\xi''$ are applied at each training iteration, the probability predictions of the same network from differently perturbed versions of the same input $x_i$ could still be different. That difference can be considered as the inconsistency error to be minimized by the consistency objective. Temporal Ensembling~\cite{laine2016temporal} utilizes the EMA of probability predictions of the student model as a teacher model's predictions, hence mitigating the high variation of teacher model predictions from one training iteration to the next. VAT~\cite{virtual-adversarial} and VAdD~\cite{adversarial-dropout} impose adversarial perturbations to either the inputs or intermediate feature vectors in directions that would potentially maximize the difference in predictions between the student model and the teacher model. 

\begin{figure*}[htb]
\centering
\mbox{
    \includegraphics[width=7.0in]{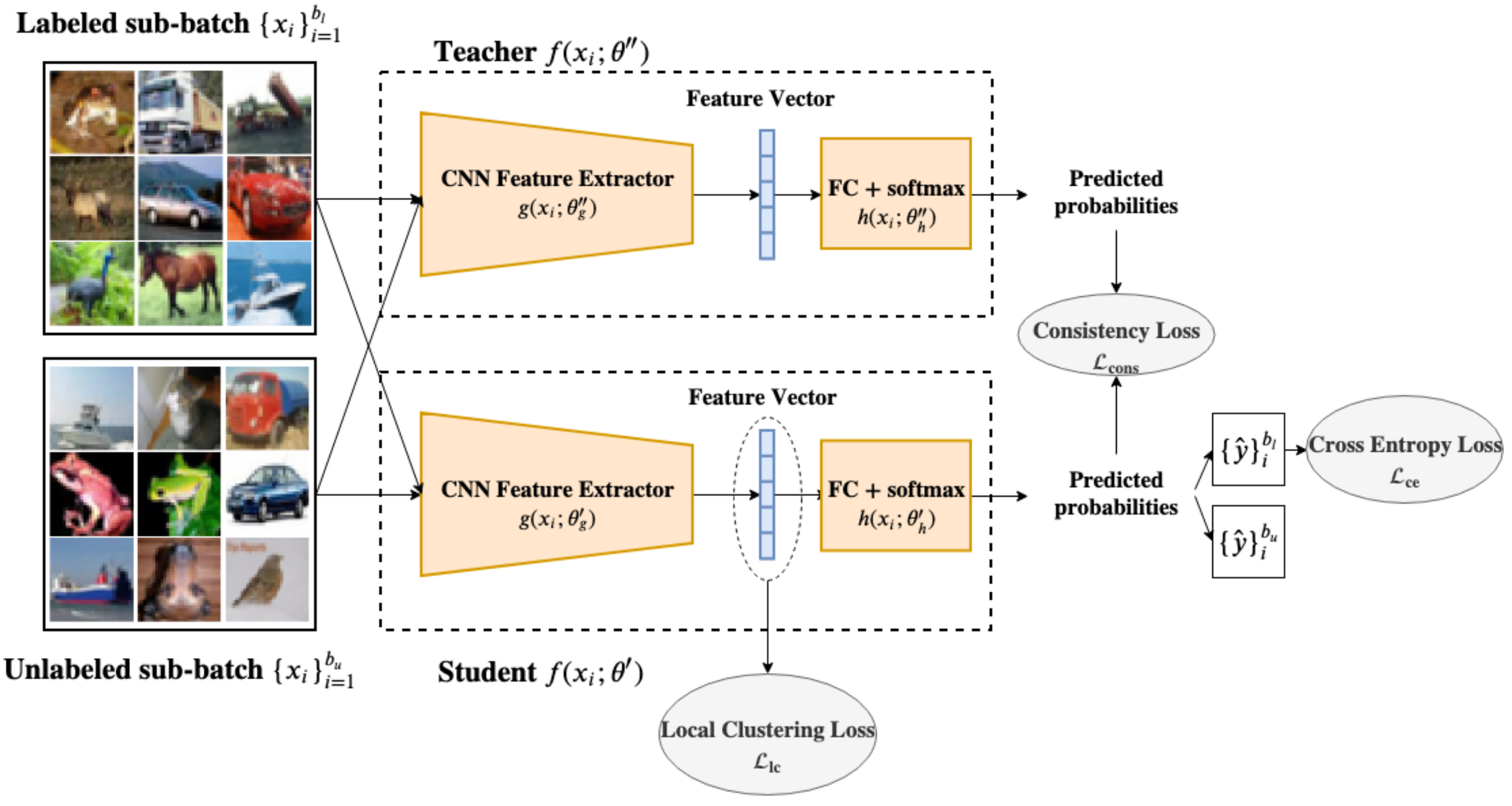}
  }
   \caption{\textbf{Network architecture.} ``FC'' represents a fully connected neural network layer.} 
\label{fig:network_architecture}
\end{figure*}

\section{Our method}
Although the Mean Teacher model performs well among SSL methods empirically, the confirmation bias~\cite{mean-teacher} issue caused by inaccurate learning objectives generated from the teacher model still exists. The incorrect learning targets can trap unlabeled data samples in low-density regions or enforce them into high-density regions of incorrect class in feature space, preventing the learning of class-specific knowledge from them. Nevertheless, this issue can be addressed if the data samples can take cues from nearby data samples in feature space.

Our idea stems from two widely used assumptions in the SSL domain:

\textit{Clustering assumption}~\cite{chapelle2003cluster}: Samples are likely to have the same class label if there is a path connecting them passing through regions of high density only.

\textit{Local consistency assumption}~\cite{zhou2004learning}: Nearby samples are likely to have the same label. Samples on the same structure (typically, a manifold) are likely to have the same label.

From \textit{local consistency assumption}, each unlabeled data point and its neighbors are likely coming from the same class. Hence, the main idea behind our method overcoming confirmation bias is to employ the help of neighboring points around every unlabeled data point to pull those misclassified unlabeled data to the high-density regions of their correct class in feature space. This is done by penalizing the magnitude of pairwise Euclidean distances between feature representations of data points and their neighbors. More specifically, we dynamically build a graph for each batch of training data (including both labeled and unlabeled data points) in feature space, and move labeled data points and their neighboring unlabeled data points closer to each other, while also moving neighboring unlabeled data points towards each other. Since labeled data are affected by class information through the supervised loss term, they implicitly lie in high-density regions of the feature space. Unlabeled data points are gradually pulled towards labeled data points of their correct class, and in consequence high-density regions, either by a direct connection or a path. Figure \ref{fig:illustrate_idea} illustrates this idea. In this context, the hope is that moving unlabeled data points towards a high-density region of their correct class would help the model discover class-specific features from unlabeled data to enable better generalization to unseen test data.

\subsection{Local Clustering}
Since our goal is to get help from neighboring samples, our question boils down to finding neighboring data samples. In traditional graph-based SSL methods \cite{zhu2003semi}, neighboring data samples are found by pre-defining an adjacency matrix in input space on some distance measure (e.g., Euclidean distance or Cosine distance), but this is infeasible for capturing the perceptual similarity between images. As an alternative, we find neighboring data samples by searching in an intermediate learned representation space. At each training iteration, we sample a sub-batch of labeled data from $\mathcal{D}_l$ of size $b_l$ and a sub-batch of unlabeled data from $\mathcal{D}_u$ of size $b_u$, and obtain their latent feature vectors by feeding them through the feature extractor network $\mathbf{z} = g(\mathbf{x}; \theta_g')$ of the student model, as shown in \ref{fig:network_architecture}. Then we perform local clustering by computing pairwise feature distances using a distance metric in feature space. In our work, we employ Euclidean distance as the distance function and empirically find it to perform well. 

More formally, we formulate pairwise distance relationships as a weighted graph in feature space, and $w_{ij}$ is the weight of an edge between samples $x_i$ and $x_j$ as 
\begin{equation}
    w_{ij} = 
    \begin{cases} exp\left(-\frac{{|| z_i - z_j ||}^2}{\epsilon}\right) & \text{\space if }|| z_i - z_j ||^2 \leq \epsilon \\
    0 & \text{ otherwise }\\
    \end{cases}
\end{equation}
where $z_i = g(x_i; \theta_g')$ and $z_j = g(x_j; \theta_g')$ are the latent feature vectors of $x_i$ and $x_j$, $\epsilon$ is a cut-off distance threshold and $||.||$ is the L2-norm. Samples $x_i$ and $x_j$ are considered as neighbors if $w_{i,j} > 0$. There are two main considerations we take into account when computing edge weights. Firstly, if two samples $x_i$ and $x_j$ are too far from each other, they are not considered as neighbors. This motivates a cut-off threshold $\epsilon$, which deems pairs with an Euclidean distance greater than $\epsilon$ as non-neighbors; secondly, the similarity between neighboring samples declines as their distance increases, so we use the negative exponential of their Euclidean distance as the distance decay function to represent their mutual effectiveness.

Similar to those deep clustering methods~\cite{hsu2015neural,Shukla2018SemiSupervisedCW}, we minimize pairwise distance for every neighboring sample pair in feature space. We formulate a local clustering loss $\mathcal{L}_{lc}$ as 
\begin{equation}
\begin{aligned}
    \mathcal{L}_{lc} = & \mathop{\mathbb{E}}_{{\{x_i\}}_{i=1}^{n_l}, {\{x_j\}}_{j=1}^{n_u}} [w_{ij} {|| g(x_i; \theta_g') - g(x_j; \theta_g') ||}^2] + \\
    & \mathop{\mathbb{E}}_{{\{x_m\}}_{m=1}^{n_u}, {\{x_n\}}_{n=1}^{n_u}} [w_{mn} {|| g(x_m; \theta_g') - g(x_n; \theta_g') ||}^2]
\end{aligned}
\end{equation}
Notice that we minimize pairwise distance between labeled samples and their neighboring unlabeled samples, as well as between neighboring unlabeled samples. It is not necessary to minimize pairwise distance between neighboring labeled samples, owing to the fact that they are already supervised sufficiently by the cross entropy loss term $\mathcal{L}_{ce}$.

Consequently, our total loss function is formulated as 
\begin{equation}
    \mathcal{L} = \mathcal{L}_{ce} + \lambda_1 \mathcal{L}_{cons} + \lambda_2 \mathcal{L}_{lc}
\end{equation}
where the coefficients $\lambda_1, \lambda_2$ are hyperparameters controlling the importance of the losses $\mathcal{L}_{cons}$ and $\mathcal{L}_{lc}$. A figure illustration of our method is presented in Figure \ref{fig:network_architecture}. Note that the computational cost introduced by local clustering regularizer is negligible since the pairwise Euclidean distances are only computed in low-dimensional feature space within a mini-batch.

It is worth pointing out the difference between our method and one other recent method, SNTG~\cite{luo2018smooth}. In SNTG, the authors are motivated to enforce smoothness across neighboring samples in feature space, while our motivation is to more directly affect the learned representations of unlabeled samples to pull them towards a high-density region of their correct class with the help of neighboring samples. SNTG constructs a teacher graph from the predicted outputs of the teacher model and uses it to guide the clustering, while our method builds a weighted graph directly from the feature representations of the student model in feature space. As we know, the predicted outputs generated by the teacher model could be inaccurate (which is why performance suffers from confirmation bias), so the clustering process in SNTG could be misguided. Moreover, we present the empirical comparison of our method versus SNTG in Section \ref{sub:sota} as well.

\section{Experiments}
In this section, we conduct extensive experiments to evaluate the performance of our proposed method and ablation studies to validate hyperparameter choices. 

\subsection{Experimental setup}
We quantitatively evaluate our proposed method on two widely used benchmark datsets: SVHN and CIFAR-10. Both datasets contain RGB images of size $32\times32$ and have 10 classes. To show that our method consistently improves the performance of Mean Teacher (MT) method~\cite{mean-teacher}, we first implement MT as the baseline, and extend it by incorporating our local clustering (LC) method. When training our LC with MT, we incorporate the LC objective after the training of MT is close to convergence. 

For a fair comparison with the state of the art methods, we employ an identical 13-layer ConvNet as used in previous works~\cite{laine2016temporal,mean-teacher,virtual-adversarial}. We train the models with a batch of 32 labeled samples and 128 unlabeled samples. We employ stochastic gradient descent (SGD) with Nesterov momentum as our optimizer. Following the practice of \cite{mean-teacher}, we too have a ramp-up phase for both consistency loss and local clustering loss where their corresponding coefficients are increased from 0 to the final values in the first few epochs right after incorporating them. The training details for each dataset are explained in the following:

\hfill \break
\noindent\textbf{SVHN}: We apply random translation to augment the training data of SVHN. When training MT model, we follow the same training schema as in MT paper~\cite{mean-teacher}, where we ramp up the consistency loss coefficient $\lambda_1$ from 0 to its maximum value 100.0 in the first 5 epochs. We adopt the same sigmoid-shaped function $e^{-5(1-x)^2}$~\cite{mean-teacher} as our ramp-up function, where $x\in[0,1]$. We set the EMA coefficient of $\alpha$ to  0.995. We train the MT model for 300 epochs with a learning rate of 0.05 and linear decay the learning rate from 0.05 to 0 in another 200 epochs afterward. When training our LC with MT, we first train MT model with the same hyperparameter settings as described above for the first 300 epochs. Then we incorporate our LC loss with a ramp-up phase of 50 epochs to increase our LC loss coefficient $\lambda_2$ from 0 to its maximum value 20.0. We employ the same ramp-up function as used in MT. We train the model with a learning rate of 0.05 for the first 400 epochs and linear decay the learning rate from 0.05 to 0 in another 200 epochs. Also, we set the cut-off threshold of $\epsilon$ to 50.0 through all the training experiments on SVHN.

\hfill \break
\noindent\textbf{CIFAR-10}: We augment CIFAR-10 training data with both random translation and horizontal flips. When training MT model on CIFAR-10, we also follow the same training schema as we train MT on SVHN. We ramp up the consistency loss coefficient $\lambda_1$ from 0 to its maximum value 100.0 in the first 5 epochs. We set the EMA coefficient of $\alpha$ to 0.99. We train the MT model for 600 epochs (on CIFAR-10 with 2,000 labeled samples) or 800 epochs (on CIFAR-10 with 4,000 labeled samples) with a learning rate of 0.05, while linearly decaying learning rates from 0.05 to 0 in the last 200 epochs afterward. When training our LC with MT, we also first train MT with the same hyperparameter settings for 600 epochs. Then we incorporate our LC loss with a ramp-up phase of 100 epochs for $\lambda_2$ from 0 to its maximum value 10.0, using the same ramp-up function as in MT. We train the model with a learning rate of 0.05 for 800 epochs (on CIFAR-10 with 2,000 labeled samples) or 1,000 epochs (on CIFAR-10 with 4,000 labeled samples) and then decay learning rate from 0.05 to 0 in the last 200 epochs afterward. Moreover, the cut-off threshold of $\epsilon$ is set to 40.0 (on CIFAR-10 with 2,000 labeled samples) or 50.0 (on CIFAR-10 with 4,000 labeled samples) through all the training experiments.

\setlength{\tabcolsep}{1pt}
\begin{table*}[htb]
\caption{Error rate percentage comparison with the state of the art methods on SVHN and CIFAR-10 over 10 runs. ``*'' indicates our re-implementation of Mean Teacher and ``LC'' denotes our local clustering method. Only methods that employ 13-layer ConvNet as their network architecture are reported for a fair comparison purpose.}
\begin{center}
\begin{tabular}{l||C{2.cm}|C{2.cm}|C{2.2cm}|C{2.2cm}}
\hline\hline
\multirow{2}{*}{Method} & \multicolumn{2}{c|}{SVHN} & \multicolumn{2}{c}{CIFAR-10} \\ \cline{2-5}
 & $n_l=500$ & $n_l=1000$ & $n_l=2,000$ & $n_l=4,000$ \\ 
\hline\hline
semi-GAN~\cite{salimans2016improved} & 18.44 $\pm$ 4.80 & 8.11 $\pm$ 1.30 & 19.61 $\pm$ 2.09 & 18.63 $\pm$ 2.32\\
Bad GAN~\cite{dai2017good} & - & 7.42 $\pm$ 0.65 & - & 14.41 $\pm$ 0.30 \\
Local GAN~\cite{qi2018global} & 5.48 $\pm$ 0.29 & 4.73 $\pm$0.29 & - & 14.23 $\pm$ 0.27 \\
$\Pi$ model~\cite{laine2016temporal} & 6.65 $\pm$ 0.53 & 4.82 $\pm$ 0.17 & - & 12.36 $\pm$ 0.31\\
TempEns~\cite{laine2016temporal} & 5.12 $\pm$ 0.13 & 4.42 $\pm$ 0.16 & - & 12.16 $\pm$ 0.31 \\
Mean Teacher~\cite{mean-teacher} & 4.18 $\pm$ 0.27 & 3.95 $\pm$ 0.19 & 15.73 $\pm$ 0.31 & 12.31 $\pm$ 0.28 \\
VAdD~\cite{adversarial-dropout} & - & 4.16 $\pm$ 0.08 & - & 11.32 $\pm$ 0.11 \\
VAT + EntMin~\cite{virtual-adversarial} & - & 3.86 $\pm$ 0.11 & - & 10.55 $\pm$ 0.05 \\
TempEns + SNTG~\cite{luo2018smooth} & 4.46 $\pm$ 0.26 & 3.98 $\pm$ 0.21 & 13.64 $\pm$ 0.32 & 10.93 $\pm$ 0.14 \\
MT + SNTG~\cite{luo2018smooth} & 3.99 $\pm$ 0.24 & 3.86 $\pm$ 0.27 & - & - \\
\hline\hline
MT* & 3.91 $\pm$ 0.11 & 3.80 $\pm$ 0.09 & 12.37 $\pm$ 0.29 & 9.93 $\pm$ 0.16\\
MT + LC \textbf{(ours)} & \textbf{3.54 $\pm$ 0.17} & \textbf{3.35 $\pm$ 0.09} & \textbf{11.56 $\pm$ 0.31} & \textbf{9.26 $\pm$ 0.16}\\
\hline
\hline
\end{tabular}
\end{center}
\label{table:sota}
\end{table*}

\subsection{Results}
\label{sub:sota}
\textbf{SVHN}: The SVHN dataset consists of 73,257 training samples and 26,032 test samples. We train the models on SVHN training images with 500 and 1,000 randomly labeled samples respectively, and evaluate the performances on the corresponding test data. We follow the same evaluation standard used in state-of-the-art approaches (comparing mean of test errors). Table \ref{table:sota} shows that incorporating our LC loss on MT improves the performance of the model (the mean MT+LC error is over 2 standard deviations lower than the mean MT error rate) in both 500 and 1,000 labeled samples settings and achieves the best test performances among state of the art methods. For a fair comparison purpose, we only compare with methods that employ the same 13-layer ConvNet as network architecture in the table. Some recent works (e.g. MixMatch\cite{berthelot2019mixmatch}, ADA-Net\cite{wang2019semi}) that adopt deep ResNet model as network architecture are not included. 

\begin{figure}[htb]
\centering
    {\includegraphics[width=0.48\linewidth]{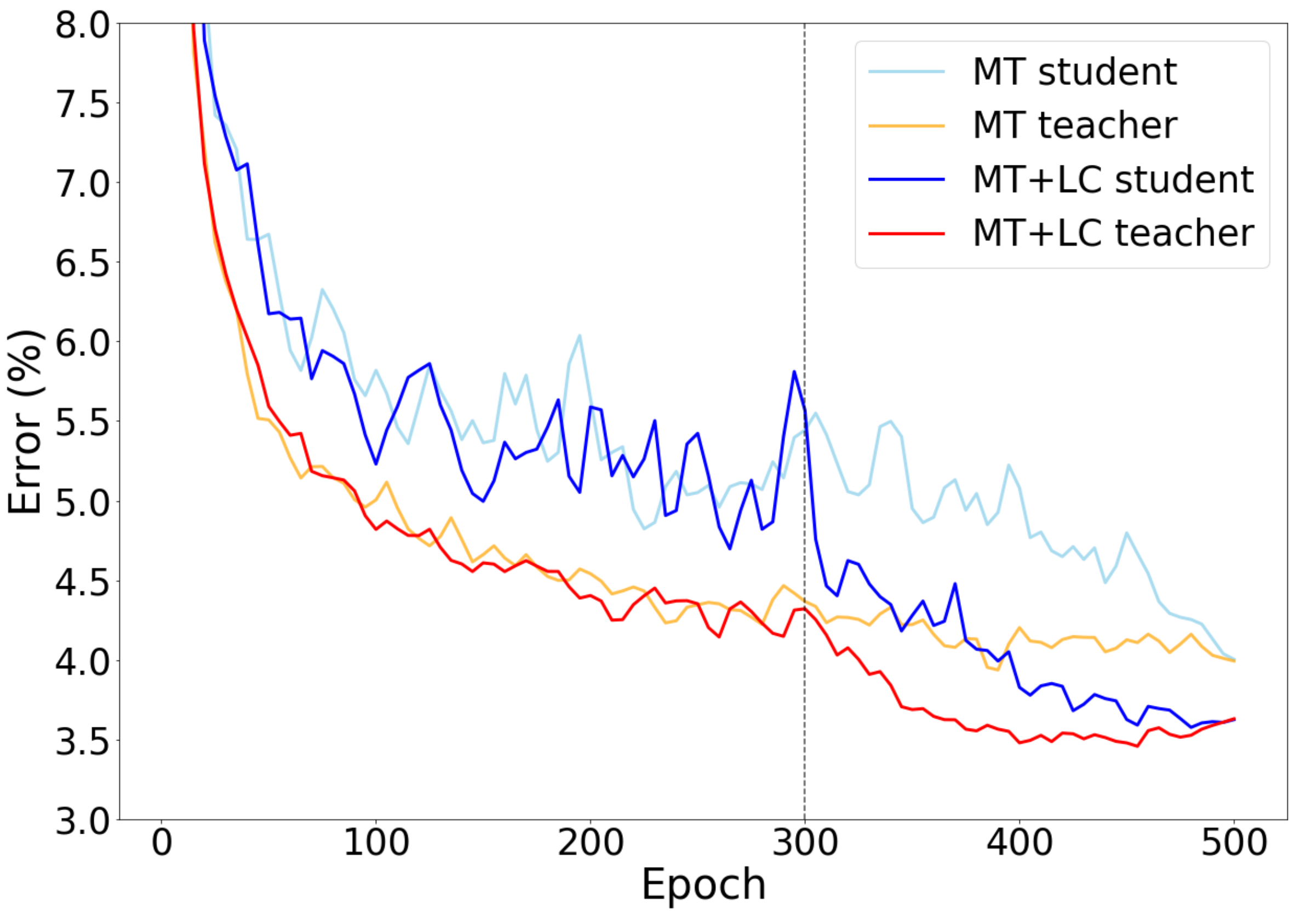} }
    {\includegraphics[width=0.48\linewidth]{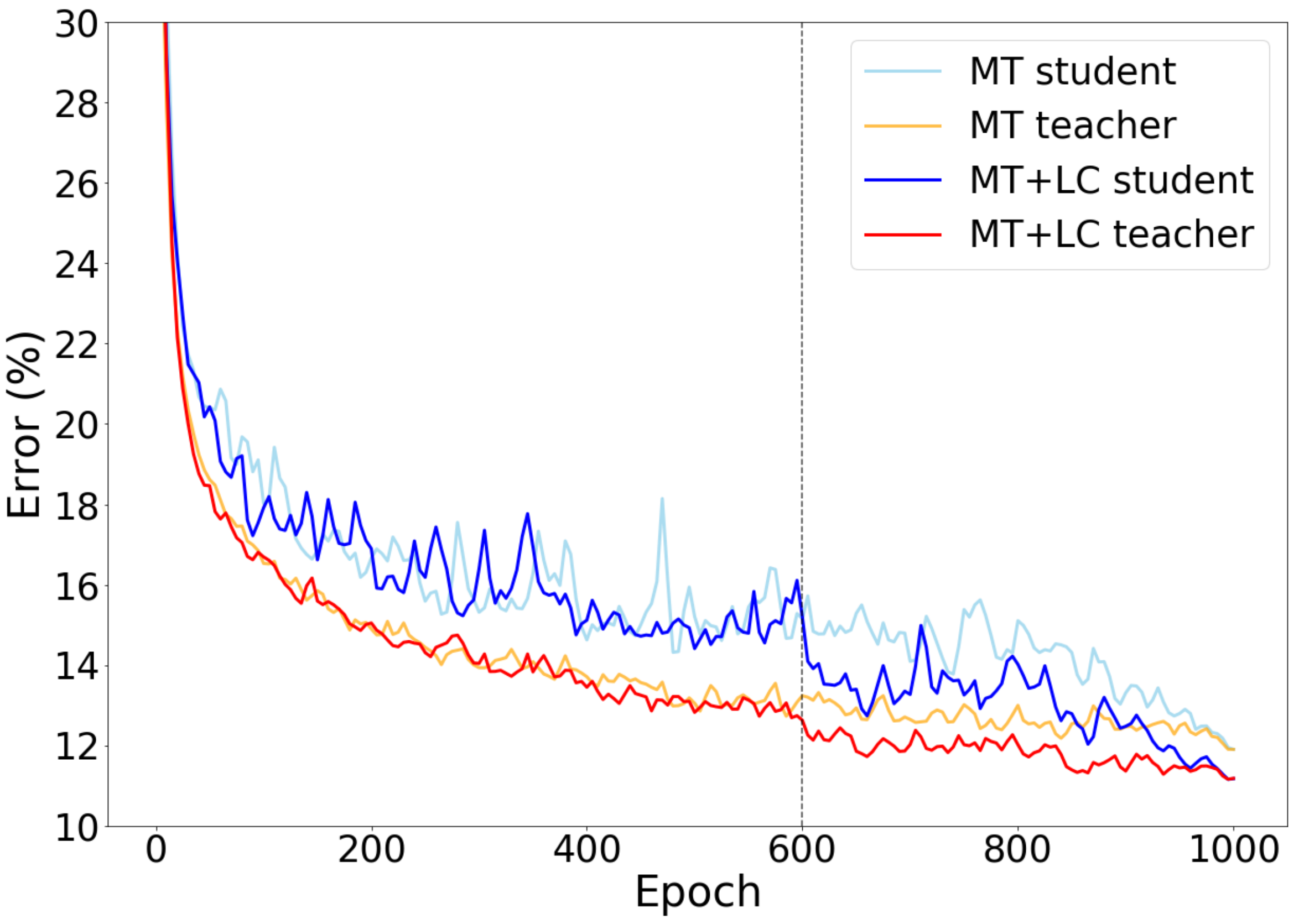} }
\caption{Smoothed test error curves of MT and MT+LC on SVHN (left) with 500 labeled samples, and CIFAR-10 (right) with 2,000 labeled samples. The LC objective is incorporated into training from 300 epochs on SVHN and 600 epochs on CIFAR-10. Best viewed in color in electronic form.}
\label{fig:error_training}
\end{figure}

Furthermore, we visualize the test error as a function of training epoch on SVHN with 500 labeled samples in Figure \ref{fig:error_training} (left). 
The figure clearly shows that there is a substantial reduction on error rate in both student and teacher models right after the LC objective is incorporated starting from 300 epochs, demonstrating that the performance improvement is owing to the introduction of LC loss. 

\hfill \break
\noindent\textbf{CIFAR-10}: The CIFAR-10 dataset consists of 50,000 training samples and 10,000 test samples. Similarly, we train the models on CIFAR-10 training images with 2,000 and 4,000 randomly labeled samples, and evaluate them on CIFAR-10 test data. As seen in Table \ref{table:sota}, LC with MT substantially improves test accuracy of the MT model, and outperforms all state of the art methods. Meanwhile, the test error curves on CIFAR-10 with 2,000 examples in Figure \ref{fig:error_training} (right) reaffirms that the improvement is consistent regardless of the dataset used.

\subsection{Ablation studies}
\label{sec:abs}
In this section, we study the effects of the two new hyperparameters introduced by our LC objective ($\epsilon$ and $\lambda_2$). Note that for these experiments, all other hyperparameters are kept fixed to values used in our implementation of the baseline MT model. We conduct all these experiments on SVHN with 500 labeled samples.

\begin{figure}[htb]
\centering
\mbox{
    \includegraphics[width=0.9\linewidth]{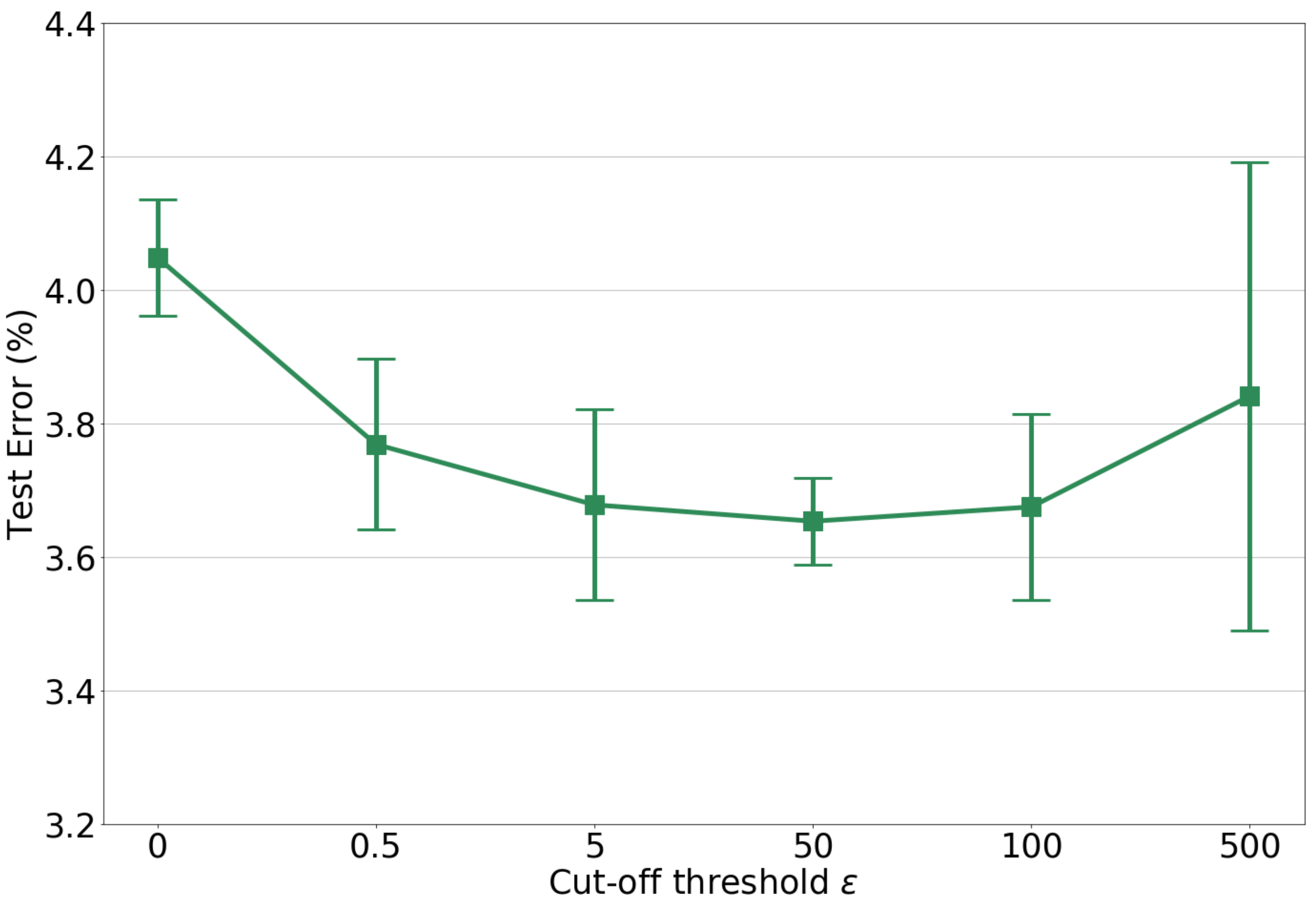}
  }
   \caption{Test errors of LC with MT with different cut-off thresholds on SVHN with 500 labeled samples over 5 runs.}
\label{fig:ab_threshold}
\end{figure}

\hfill \break
\textbf{Effect of cut-off threshold $\epsilon$ and }: One essential hyperparameter introduced by our LC loss is the cut-off threshold $\epsilon$. If $\epsilon$ is too small, almost no neighbors are found around each data point. To the contrary, distant points are considered as neighbors when $\epsilon$ is too large. We vary $\epsilon$ in a broad range of values and the results are shown in Figure \ref{fig:ab_threshold}. Note that the model with $\epsilon=0$ is equivalent to the MT model as no neighbors would be considered in this scenario. From the figure, we observe that there is a decreasing trend on error rate when increasing $\epsilon$ from 0 to 50, which implies LC works better with more neighbors taken into account. However, there is an error rate increase if $\epsilon$ is increased further, indicating too large $\epsilon$ would degrade model performance. In our experiments, we also find that the LC method would fail for $\epsilon > 500$, leading to a scenario we term ``distribution collapse'', where the entire dataset is mapped to a single point by the feature extractor network. Nevertheless, all explored values of $\epsilon$ in a wide range that do not lead to distribution collapse improve upon the MT model.  

\begin{figure}[htb]
\centering
\mbox{
    \includegraphics[width=0.9\linewidth]{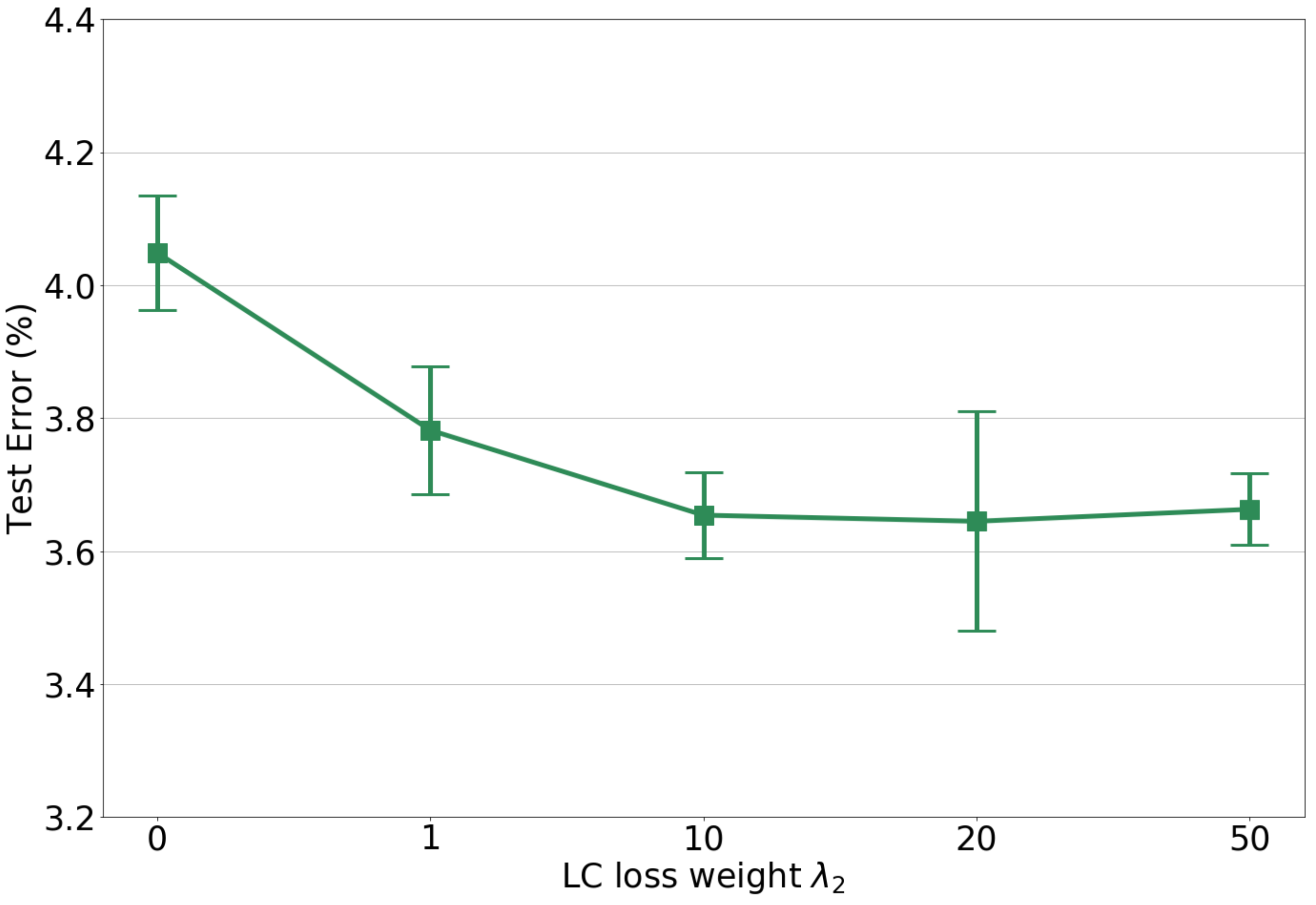}
  }
   \caption{Test errors of LC with MT with different LC loss weights on SVHN with 500 labeled samples over 5 runs.}
\label{fig:ab_weight}
\end{figure}

\hfill \break
\textbf{Effect of LC loss weight $\lambda_2$}: We also evaluate model performance on different $\lambda_2$ values, as shown in Figure \ref{fig:ab_weight}. Similarly, the model with $\lambda_2=0$ is purely a MT model. The figure shows that the error rate decreases gradually as $\lambda_2$ increases, and arrives at a steady state when $\lambda_2 \ge 10$. The model sees a distribution collapse failure case similar to $\epsilon > 500$ when $\lambda_2 > 50$, owing to the fact that LC loss dominates the learning process. Similarly, all explored values of $\lambda_2$ in a wide range that do not lead to distribution collapse improve upon the MT model. 

\begin{figure}[htb]
\begin{subfigure}{0.48\linewidth}
\centering
    {\includegraphics[width=\linewidth]{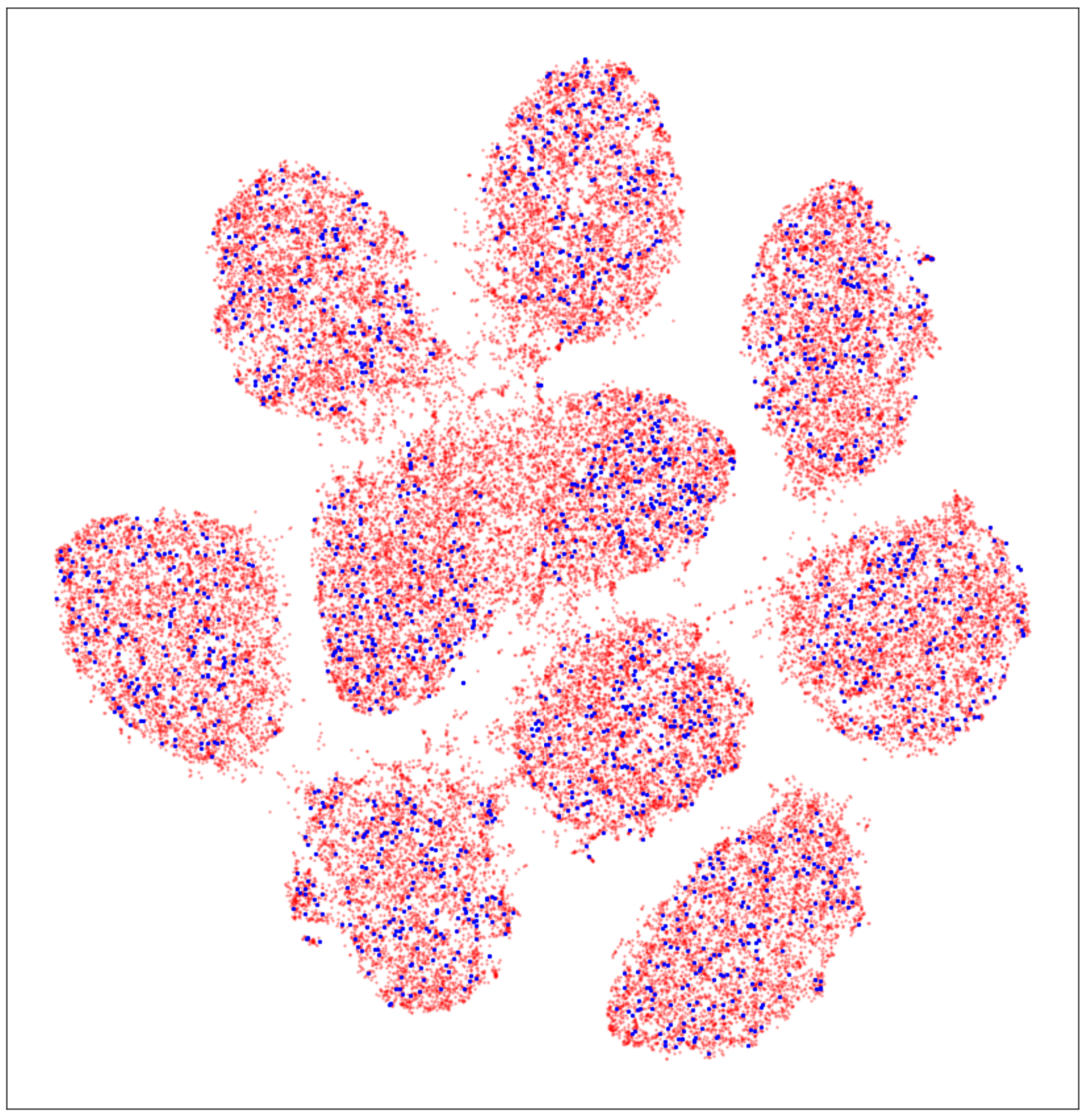} }
    \caption{MT, Training Data}
\end{subfigure}
\hfill
\begin{subfigure}{0.48\linewidth}
\centering
    {\includegraphics[width=\linewidth]{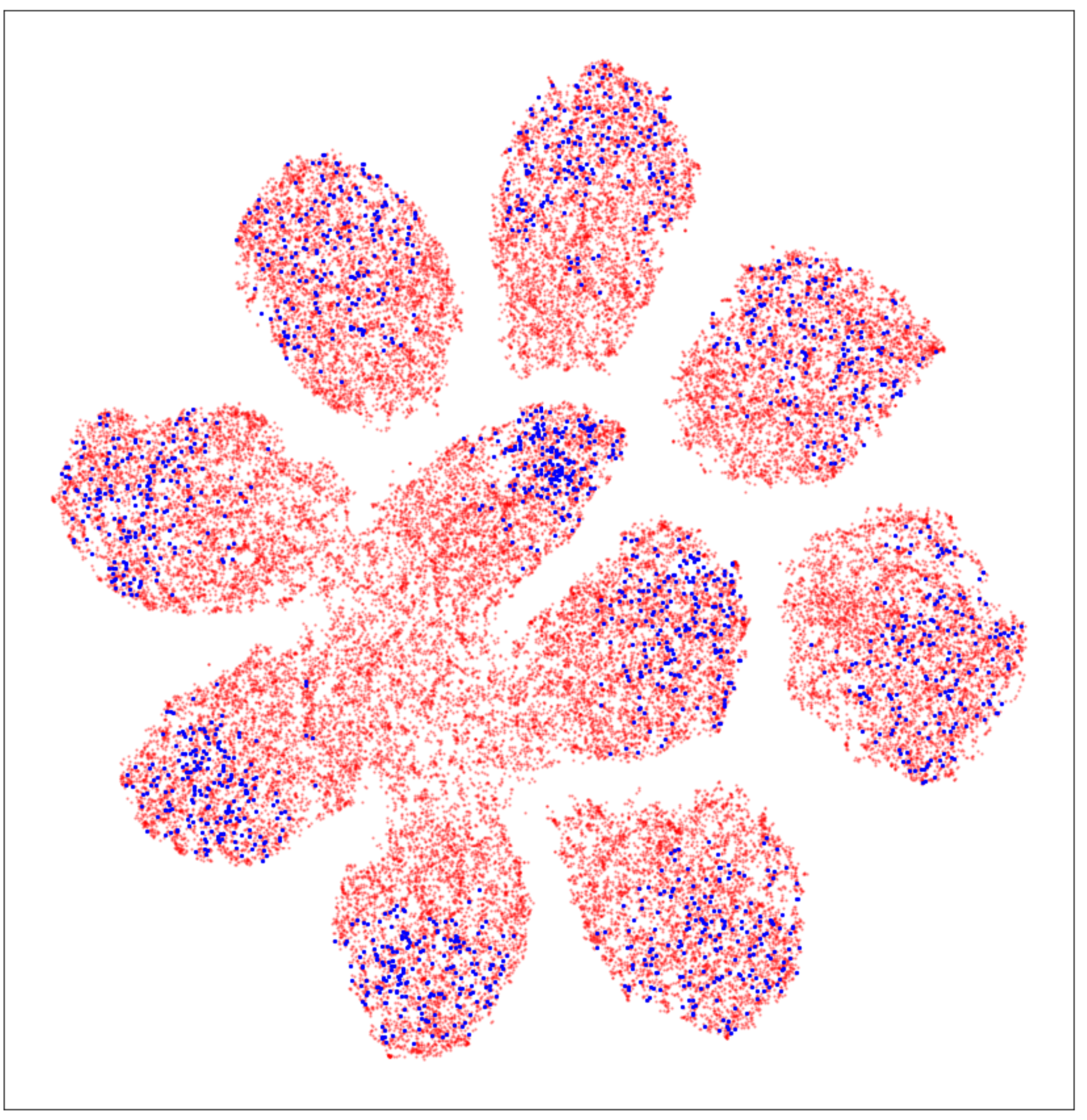} }
    \caption{MT + LC, Training Data}
\end{subfigure}
\caption{t-SNE Visualization of CIFAR-10 training data features obtained by MT (left) and MT + LC (right). The models are trained on CIFAR-10 with 2,000 labeled samples. Unlabeled samples are in red and labeled samples are in blue. Best viewed in color in electronic form.}
\label{fig:vis_train}
\end{figure}

\subsection{Visualization}
To better understand the effect of LC loss, we visualize the intermediate learned representations of MT and MT + LC on CIFAR-10 training data with t-Distributed Stochastic Neighbor Embedding (t-SNE~\cite{maaten2008visualizing}) in Figure \ref{fig:vis_train}. We train the models on CIFAR-10 with 2,000 labeled samples, extract and project the intermediate learned representations from the layer on which LC loss was added $\boldsymbol{z} \in \mathbb{R}^{128}$ for all training data into 2-dimensional feature space.  From the figure, we observe that the labeled data of the different blobs (which also correspond roughly to predicted class labels) are more separated in MT + LC (right) compared to MT (left). Additionally, LC causes the space to warp such that the labeled points take a majority of unlabeled points of the same class to a high-density region that hugs the edge of its cluster. An additional insight here is that this visualization shows counter to intuition that distribution alignment between labeled and unlabeled points is not necessary for higher classification accuracy.



\begin{figure}[htb]
\begin{subfigure}{0.48\linewidth}
\centering
    {\includegraphics[width=\linewidth]{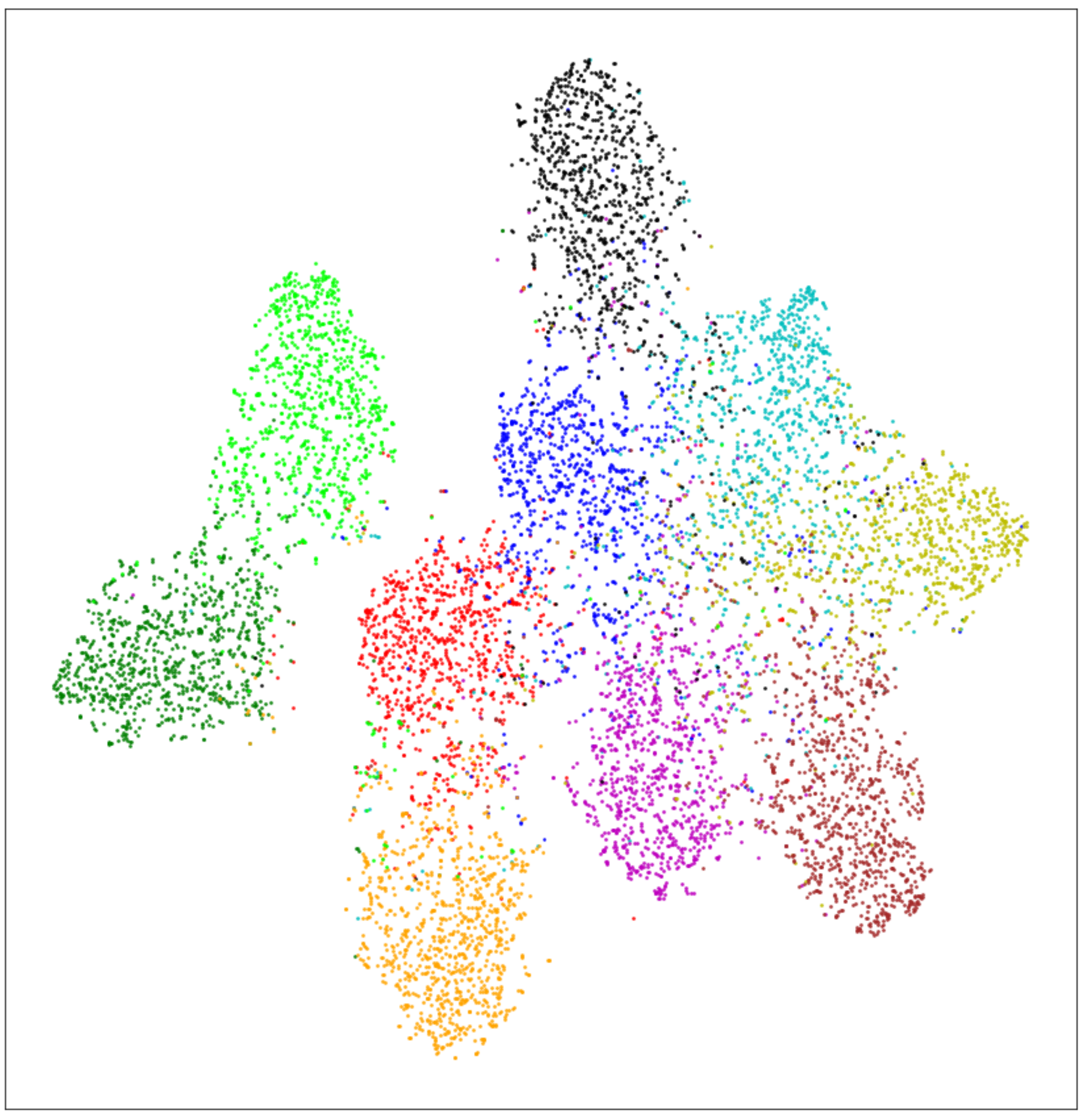} }
    \caption{MT, Test Data}
\end{subfigure}
\hfill
\begin{subfigure}{0.48\linewidth}
\centering
    {\includegraphics[width=\linewidth]{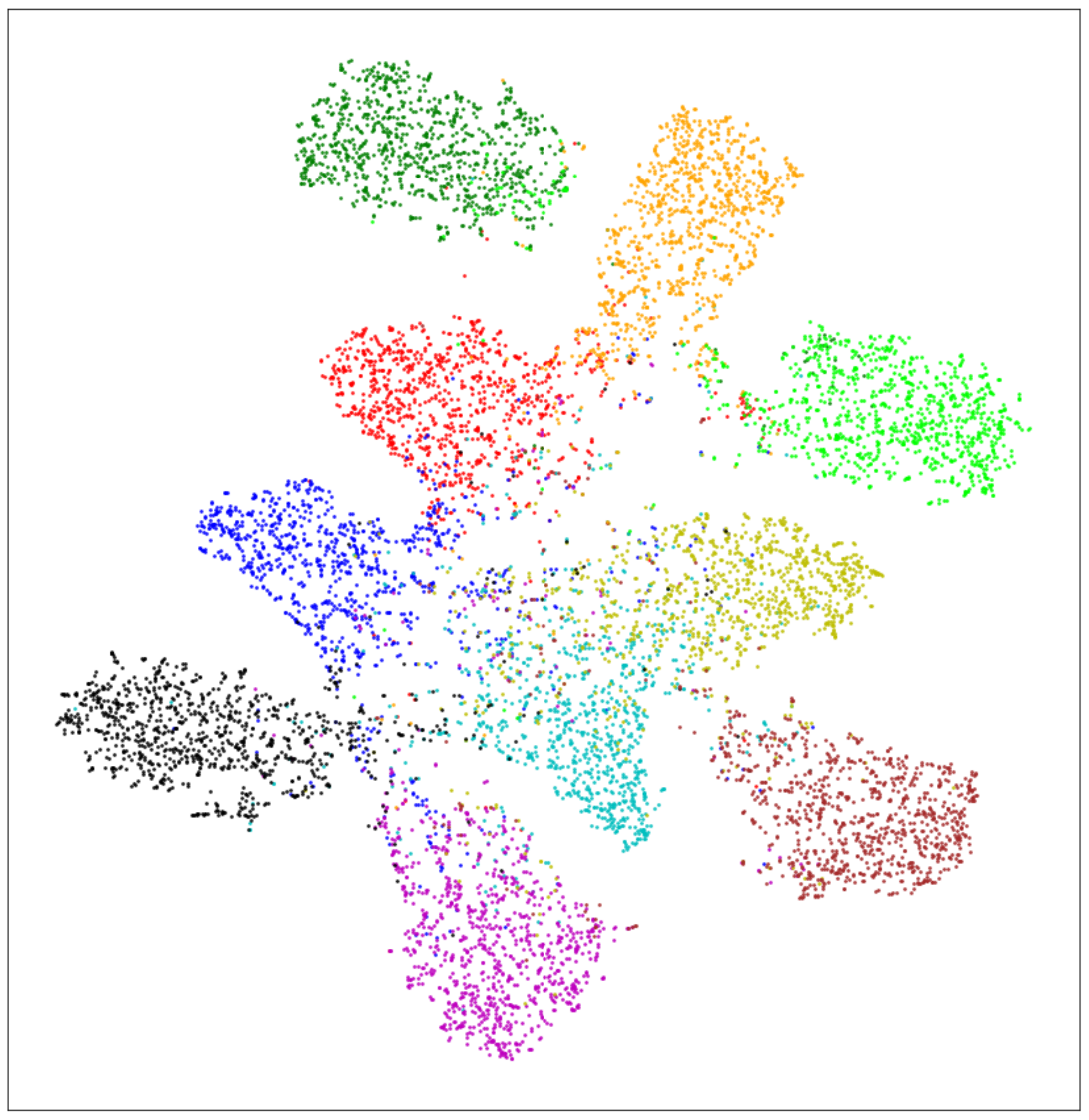} }
    \caption{MT + LC, Test Data}
\end{subfigure}
\caption{t-SNE Visualization of CIFAR-10 test data features obtained by MT (left) and MT + LC (right). The models are trained on CIFAR-10 with 2,000 labeled samples. Each color denotes a ground truth class. Best viewed in color in electronic form.}
\label{fig:vis_test}
\end{figure}

We also visualize the intermediate learned representations of MT and MT + LC on CIFAR-10 test data with t-SNE in Figure \ref{fig:vis_test}. From the figure, it is undeniable that the learned representations of different classes are more distinctively separated in MT + LC (right), while they are mixed in MT (left), validating that the LC loss helps learn more class-specific knowledge.

\section{Conclusions and Future Work}
\label{sec:conclusion}
In this work, we have proposed a new method called Local Clustering to tackle the confirmation bias issue in Mean Teacher method. In particular, it considers the correlations between nearby data points in feature space and hence corrects misclassified data points by pushing them to the high-density region of their ground truth class with the help of neighboring points. In our experiments, we validate the effectiveness of our method with MT on two benchmark datasets SVHN and CIFAR-10, and achieve performance comparable to state of the art among semi-supervised methods.

Even though we only focus on the application of our LC method on MT to semi-supervised classification in this work, the LC loss can be regarded as a general regularization technique that might provide an interesting foundation for use in other applications and existing methods.

\bibliographystyle{IEEEtran}
\bibliography{ieee}

\end{document}